\documentclass[sigconf]{acmart}

\AtBeginDocument{%
  \providecommand\BibTeX{{%
    \normalfont B\kern-0.5em{\scshape i\kern-0.25em b}\kern-0.8em\TeX}}}

\usepackage[ruled, vlined, linesnumbered]{algorithm2e}
\usepackage{subcaption}
\usepackage{tabularx}
\usepackage{multirow}
\begin{document}

\title{Improving Camouflaged Object Detection with the Uncertainty of Pseudo-edge Labels}

\author{Nobukatsu Kajiura}
\email{kajiura@nii.ac.jp}
\affiliation{
  \institution{The University of Tokyo}
  \institution{National Institute of Informatics}
  \state{Tokyo}
  \country{Japan}}

\author{Hong Liu}
\email{hliu@nii.ac.jp}
\affiliation{
  \institution{National Institute of Informatics}
  \state{Tokyo}
  \country{Japan}}

\author{Shin’ichi Satoh}
\email{satoh@nii.ac.jp}
\affiliation{
  \institution{National Institute of Informatics}
  \institution{The University of Tokyo}
  \state{Tokyo}
  \country{Japan}}

\begin{abstract}
    This paper focuses on camouflaged object detection (COD), which is a task to detect objects hidden in the background.
    Most of the current COD models aim to highlight the target object directly while outputting ambiguous camouflaged boundaries. 
    On the other hand, the performance of the models considering edge information is not yet satisfactory.
    To this end, we propose a new framework that makes full use of multiple visual cues, \emph{i.e.}, saliency as well as edges, to refine the predicted camouflaged map.
    This framework consists of three key components, \emph{i.e.}, a pseudo-edge generator, a pseudo-map generator, and an uncertainty-aware refinement module.
    In particular, the pseudo-edge generator estimates the boundary that outputs the pseudo-edge label, and the conventional COD method serves as the pseudo-map generator that outputs the pseudo-map label. Then, we propose an uncertainty-based module to reduce the uncertainty and noise of such two pseudo labels, which takes both pseudo labels as input and outputs an edge-accurate camouflaged map.
    Experiments on various COD datasets demonstrate the effectiveness of our method with superior performance to the existing state-of-the-art methods.
\end{abstract}




\maketitle

\section{Introduction}

\begin{figure}[!t]
  \begin{minipage}[b]{0.19\hsize}
    \centering
    \includegraphics[width=\hsize]{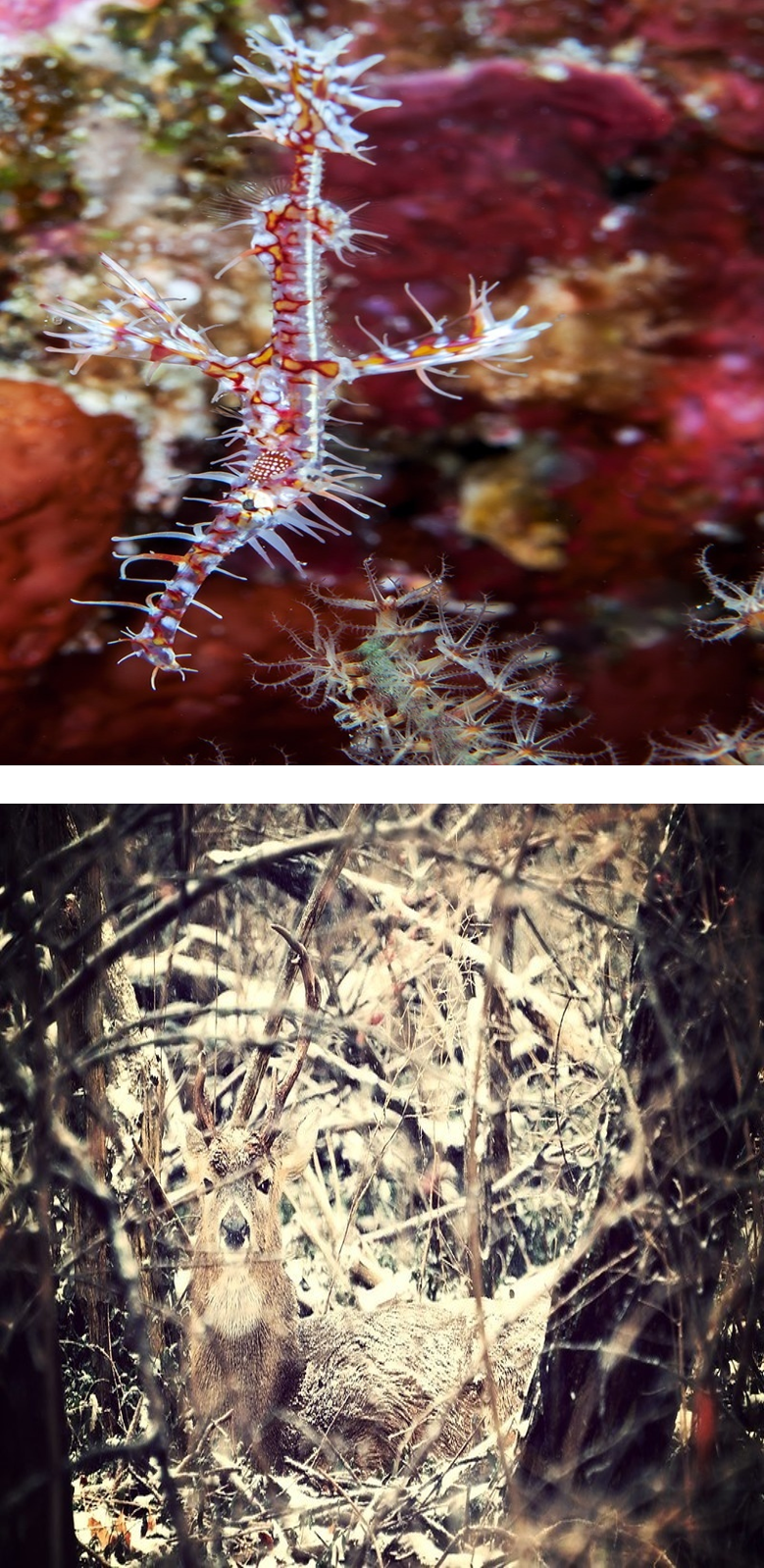}
    \subcaption{Image}
  \end{minipage}
  \begin{minipage}[b]{0.19\hsize}
    \centering
    \includegraphics[width=\hsize]{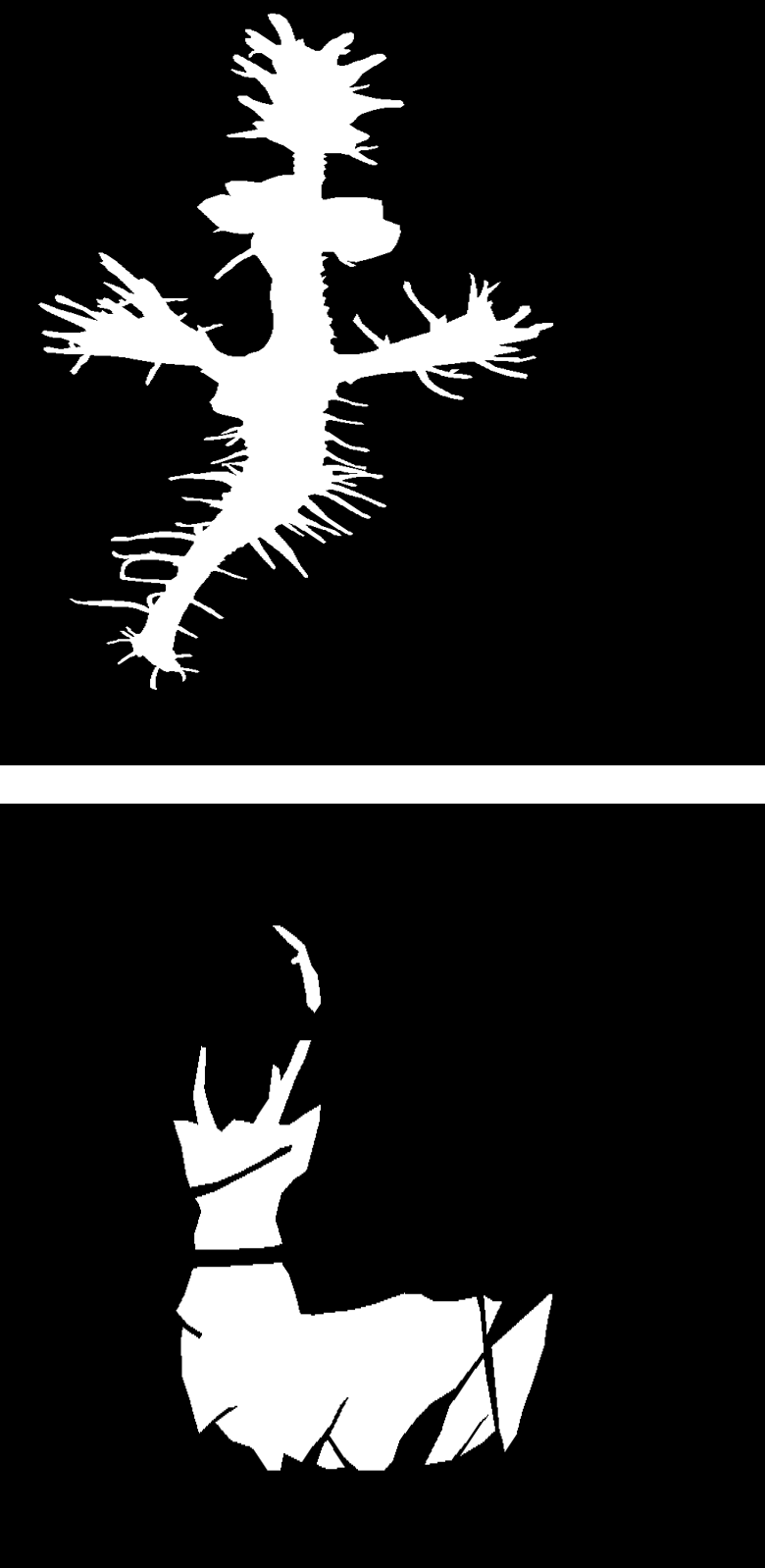}
    \subcaption{GT}
  \end{minipage}
  \begin{minipage}[b]{0.19\hsize}
    \centering
    \includegraphics[width=\hsize]{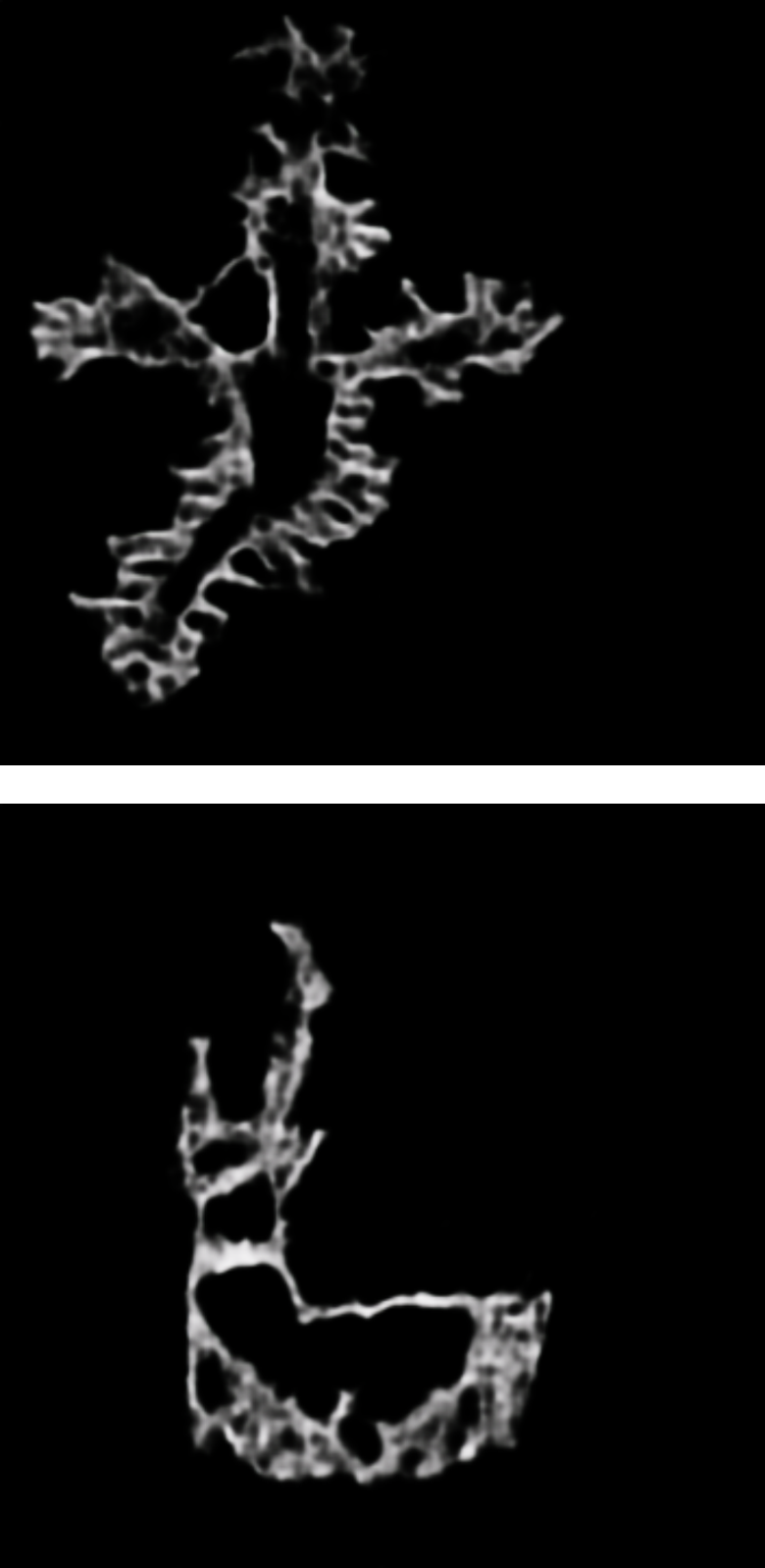}
    \subcaption{P-Edge}
  \end{minipage}
  \begin{minipage}[b]{0.19\hsize}
    \centering
    \includegraphics[width=\hsize]{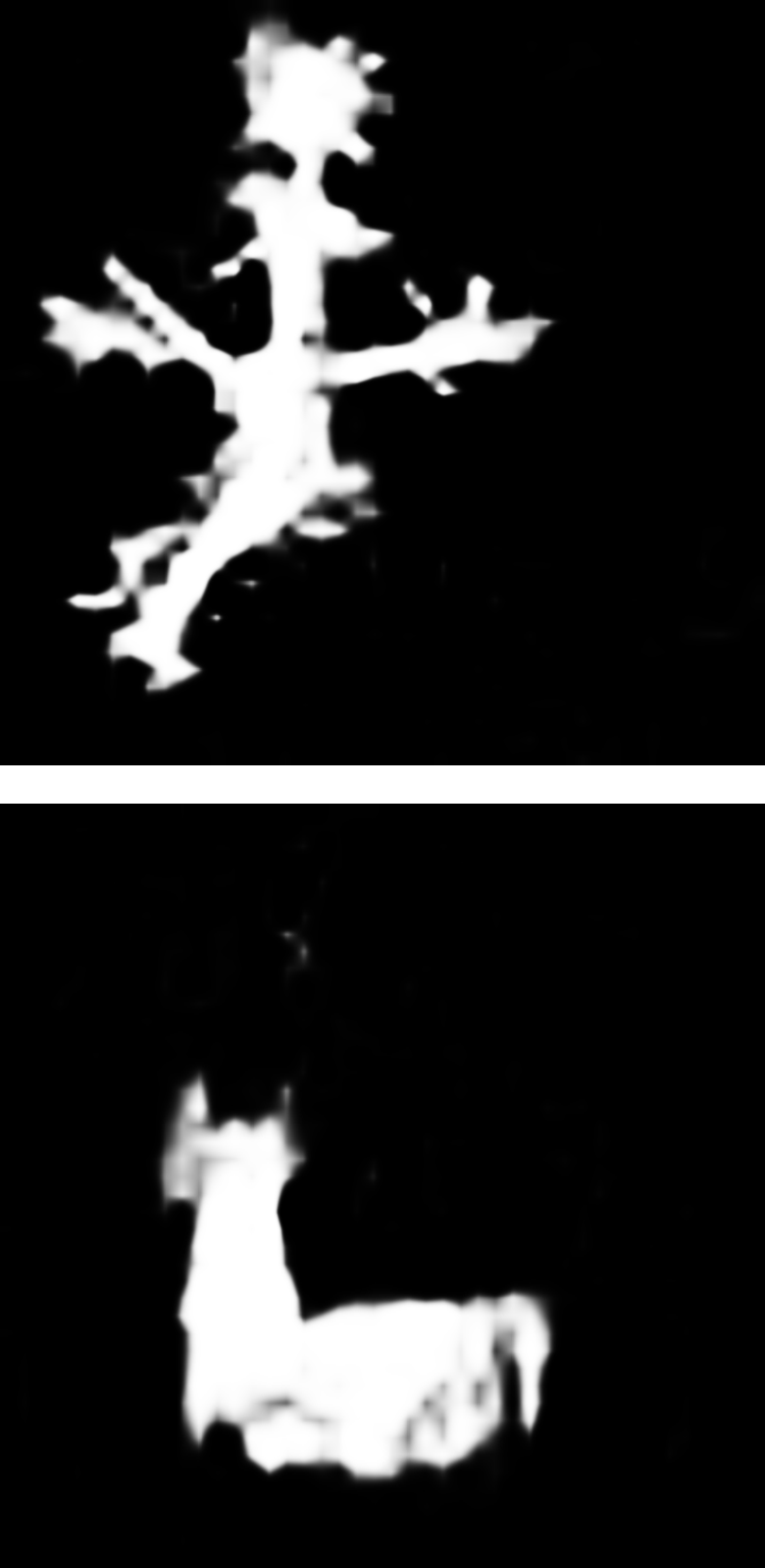}
    \subcaption{P-Map}
  \end{minipage}
  \begin{minipage}[b]{0.19\hsize}
    \centering
    \includegraphics[width=\hsize]{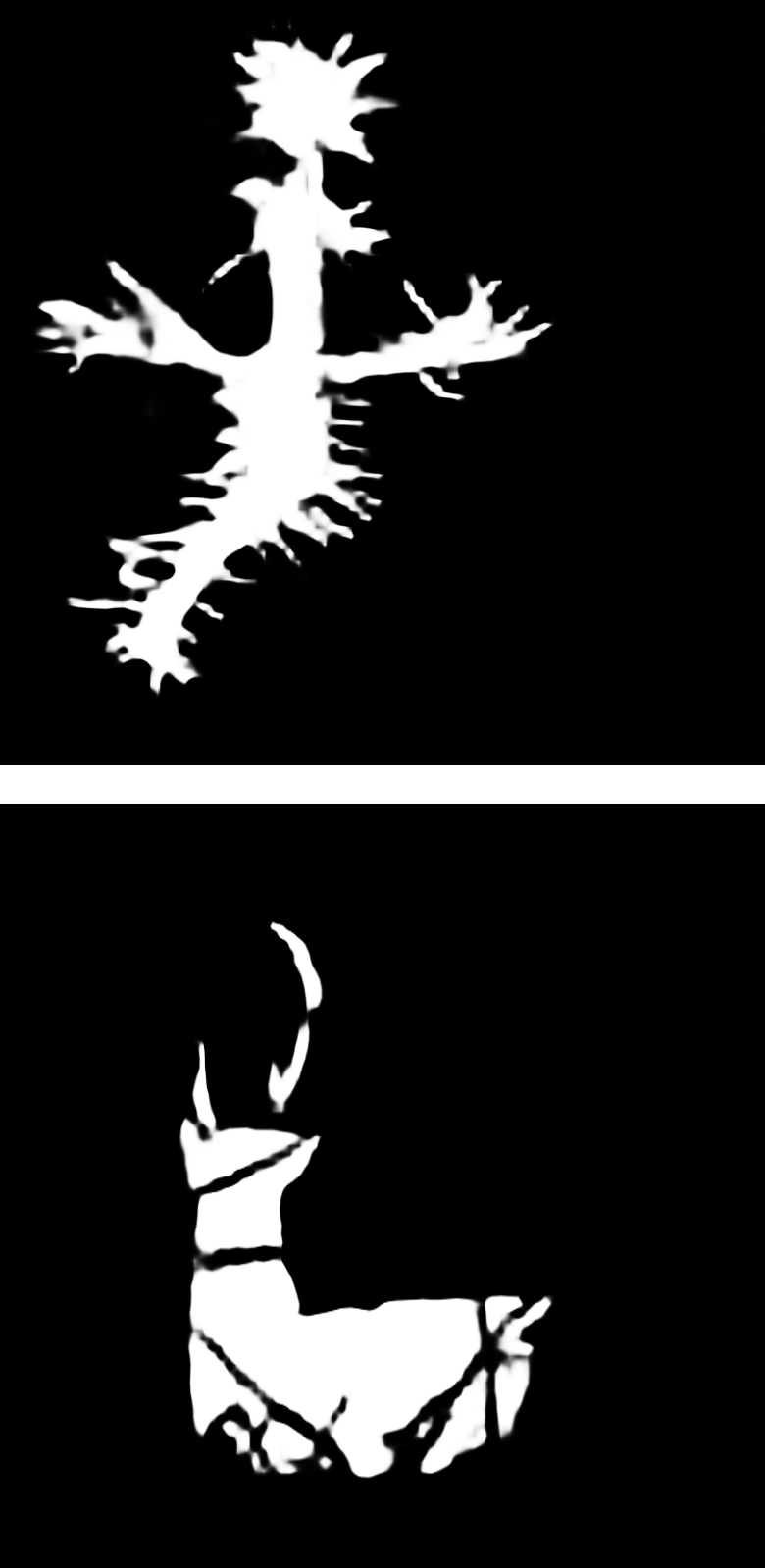}
    \subcaption{Ours}
  \end{minipage}\vspace{-1em}
  \caption{Illustration of the proposed method. "P-" represents a pseudo label, and our method outputs edge-accurate camouflaged maps based on the uncertain pseudo-map labels and pseudo-edge labels. Note that any conventional COD method can be used as the pseudo-map generator~(PMG), and state-of-the-art SINet-v2~\cite{fan2021concealed} is used here.}
  \label{fig:our_method}
\end{figure}

Camouflage is a way for creatures in nature to blend into the surroundings to make it harder for preys and predators to notice themselves~\cite{cuthill2005disruptive}.
Camouflaged Object Detection (COD) is a task of detecting objects from images containing such camouflaged objects~\cite{fan2020camouflaged}, as shown in the first column of Figure \ref{fig:our_method}.
COD is expected to have a wide range of applications, including ecological protection, medicine, surveillance systems, search and rescue systems for disaster, military applications, anomaly detection~\cite{fan2020camouflaged}.
In addition, it is expected that improving the accuracy of COD would improve the accuracy of generic object detection.

COD is a more challenging task compared to generic object detection~\cite{liu2020deep}.
This is because the foreground has a similar texture to the background, which is not easy for a human to notice the camouflaged object~\cite{stevens2009animal}.
In existing COD methods \cite{fan2020camouflaged,fan2021concealed}, due to the boundaries of camouflaged objects are not clear enough, they often result in blurred and ambiguous boundaries of the output camouflaged map. 
Inspired by \cite{osorio1991camouflage,egan2016edge}, making good use of edge information is helpful for performance improvement. However, the performance of recent models \cite{zhai2021Mutual,sun2021context}, that consider edge information, achieve lower performance than regular COD models like \cite{fan2021concealed}.
Therefore, we aim to integrate the edge information into the process of COD, which is to refine the predicted camouflaged maps by referring to the separately predicted edges.

In this paper, we propose a new framework, called \textbf{Uncertainty Reduction COD}~(UR-COD), that explicitly considers the edge information of the camouflaged object, which is to enhance the current COD model by introducing a camouflaged edge detection module and to output camouflaged map with clear defined boundary~(as shown in Figure~\ref{fig:our_method}).
The proposed framework can extract powerful features that identify more details of camouflaged objects than models that do not or implicitly consider the boundaries of camouflaged objects.
In particular, we first introduce a camouflaged edge detection module, in which output edge is used as a pseudo-edge label.
Second, we use a conventional COD model to generate the coarse camouflaged map that can be seen as a pseudo-map label.
Consequently, our goal is to generate an accurate camouflaged map by given the pseudo-edge label and pseudo-map label.
However, these pseudo labels are noisy and contain uncertainty compared to ground-truth labels.  
To this end, we use the Conditional Variational Auto-encoder (CVAE)~\cite{sohn2015learning} to build an uncertainty-aware map refinement module, which can output a camouflaged map even in the presence of noisy pseudo label inputs.
It is worth noting that our framework can use any existing COD models as the pseudo-map generator, and our proposed uncertainty-enhanced model consistently improves the performance of the original model. 
The whole framework of our proposed model is shown in Figure \ref{fig:our_framework}. 

To verify the usefulness of our method, quantitative evaluation is conducted on the CAMO dataset~\cite{le2019anabranch}, CHAMELEON dataset~\cite{skurowski2018animal}, COD10K dataset~\cite{fan2020camouflaged}, and NC4K dataset~\cite{yunqiu_cod21}.
The results demonstrate that our framework outperforms the corresponding conventional COD methods in almost all of the four widely-used evaluation metrics, which are S-measure~\cite{fan2017structure}, E-measure~\cite{fan2018enhanced}, weighted F-measure~\cite{margolin2014evaluate}, and MAE.
Furthermore, it is shown that the best performing COD model exceeded the performance of the state-of-the-art model when used as a pseudo-map generator.

Our main contributions are summarized as follows:
\begin{itemize}
  \item We explicitly consider uncertain camouflaged edge and camouflaged map information and improve the conventional COD methods.

  \item Experiments on various COD datasets confirm that the performance of our method outperforms that of the conventional state-of-the-art methods.
\end{itemize}

\section{Related Work}
\subsection{Camouflaged Object Detection}
Previous COD methods can be categorized into either low-level-features-based methods or deep-features-based methods.
Low-level-features-based methods are methods for detecting camouflaged objects based on low-level features, such as color, shape, and brightness of the image.
There are many methods for this research~\cite{sengottuvelan2008performance, siricharoen2010robust,kavitha2011efficient,liu2012foreground,song2010new,pan2011study,hou2011detection,gallego2014foreground}, but their performance is not yet satisfactory.
This is because a well-performed camouflage is good at deceiving low-level features, so it is difficult to detect camouflaged objects using low-level features.
To solve this problem, deep-features-based methods have been attracting attention in recent years.

Deep feature-based methods use deep neural networks to correctly output the camouflaged maps from an input image during training, and predict the camouflaged maps of the objects during testing.
Le et al.~\cite{le2019anabranch} proposed ANet that addressed both classification and segmentation tasks.
Fan et al.~\cite{fan2020camouflaged, fan2021concealed} introduced SINet and SINet-v2 that first roughly searched for camouflaged objects and then identified their segments.
Yan et al.~\cite{yan2020mirrornet} proposed MirrorNet that focused on instance segmentation and adversarial attack.
Sun et al.~\cite{sun2021context} proposed C$^2$FNet, which integrated the cross-level features with the consideration of rich global context information.
Mei et al.~\cite{mei2021camouflaged} proposed PFNet via the distraction mining strategy.
Lyu et al.~\cite{yunqiu_cod21} introduced camouflaged object discriminative region localization and camouflaged object ranking.
Zhang et al.~\cite{zhang2021depth} introduced RGB-D COD with a monocular depth estimation network.
According to the current leaderboard\footnote{http://dpfan.net/Camouflage/}, SINet-v2 has shown particularly high performance that directly segments objects, while the boundaries of the output camouflaged map are often ambiguous.

On the other hand, some methods focus on edge information of camouflaged objects.
Wang et al.~\cite{wang2019salient} used edge information to improve the performance of salient object detection.
Zhu et al.~\cite{zhu2021inferring} proposed TINet, which interactively refined multi-level texture including contour edge and Canny edge~\cite{canny1986computational}.
Zhai et al.~\cite{zhai2021Mutual} introduced a graph-based joint learning framework for detecting camouflaged objects and edges.

Though these deep-features-based methods show considerably higher performance than low-level-features-based methods, methods that segment objects directly have the problem of edge clarity, while methods that consider edges have the problem of accuracy.
Therefore, to overcome these problems, we aim to combine the best of both methods and propose to refine the directly predicted camouflaged maps by referring to the separately predicted edges.
Moreover, to solve the problem of the low performance of edge-based methods, our method takes into account the uncertainty of the predicted edges.

\begin{figure*}[tb]
	\begin{center}
		\includegraphics[width=0.8\linewidth]{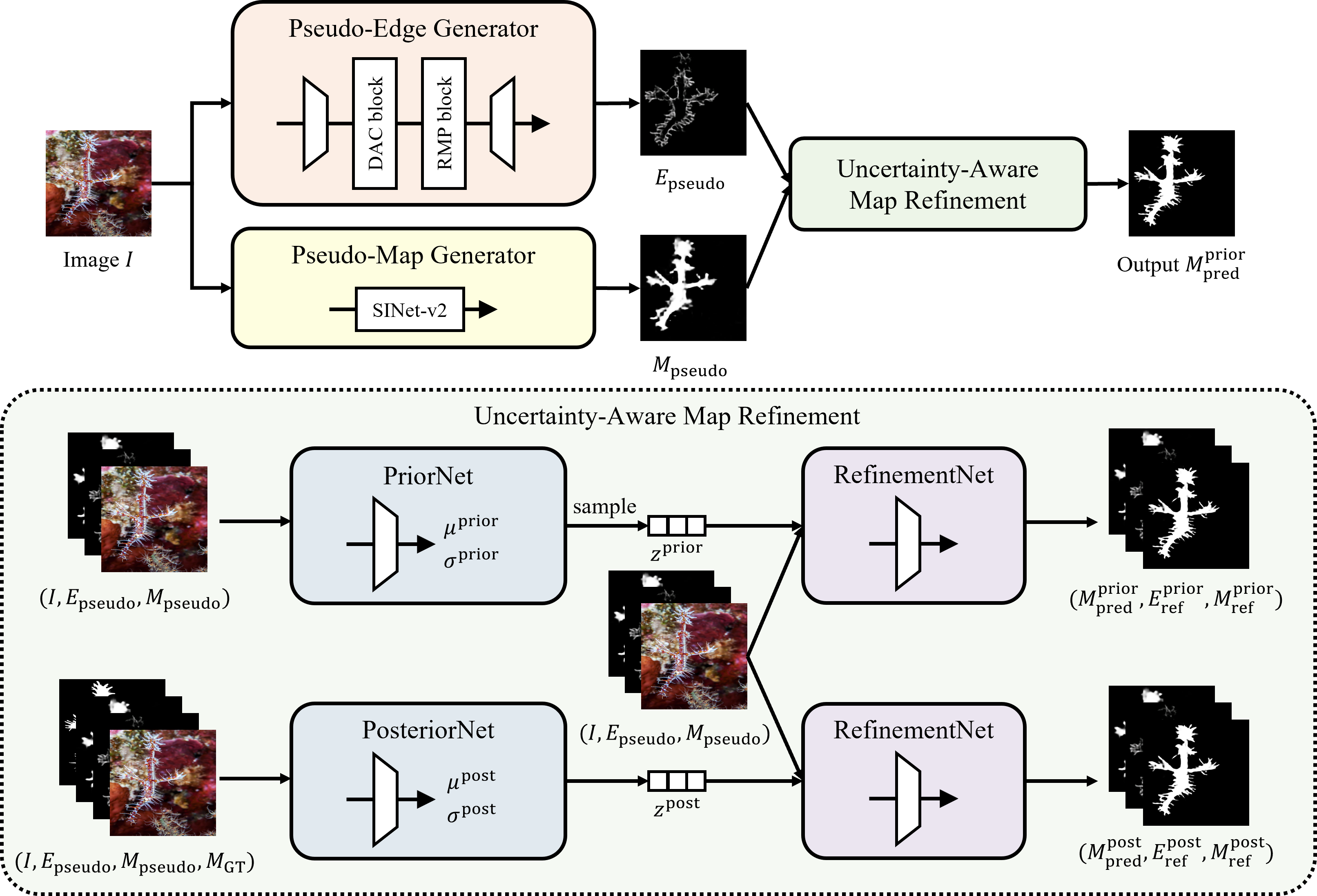} \vspace{-0.5em}
		\caption{Overview of our framework. The obscure boundaries of pseudo-map labels generated by the pseudo-map generator~(PMG) are refined by the uncertainty-aware map refinement~(UAMR) module using pseudo-edge labels generated by the pseudo-edge generator~(PEG).}
		\label{fig:our_framework}
	\end{center}
\end{figure*}

\subsection{Edge detection}
With the development of deep learning, many high-performance edge detection methods have been proposed~\cite{xie2015holistically,liu2017richer,bertasius2015deepedge,soria2020dexined}.
However, most of the objects in the datasets handled by these methods are general salient objects, and they do not provide sufficient performance for camouflaged objects.
Gu et al.~\cite{gu2019net} proposed CENet that uses a UNet-based model for improving the medical image segmentation.
This method extracts semantic information of the context and generates a high-level feature map by using dense atrous convolution~(DAC) blocks~\cite{gu2019net} and residual multi-kernel pooling~(RMP) blocks~\cite{gu2019net}.
Therefore, compared to other methods that do not consider semantic information, this method is likely to be applicable even when the detection target is not salient.

\subsection{Uncertainty-aware object detection}
Existing methods for object detection treat saliency map prediction as a point estimation problem by learning the correspondence between input images and ground-truth maps, which deterministically generate a single saliency map.
However, since ground-truth map labels are based on human annotations, especially in the field of salient object detection, the recognition of the most salient objects differs among annotation generators, and the ground-truth labels themselves contain uncertainty.
Zhang et al.~\cite{zhang2020uc} solved this problem of label uncertainty by using a Conditional Variational Auto-encoder~(CVAE)~\cite{sohn2015learning}, that is called UCNet.
UCNet introduces CVAE to model uncertainty of human annotation, and accurate object detection is achieved.
In our proposed method, we generate pseudo labels for camouflaged maps and edges to obtain accurate COD.
Since the pseudo labels are generated by a learning-based method, there is some uncertainty in the pseudo labels.
To address this problem, it is necessary to consider the uncertainty.

\section{Method}
The purpose of this study is to predict camouflaged maps with well-defined boundaries from an input image containing camouflaged objects.
In conventional COD methods, there lacks a model that takes the edge information into account while the camouflaged map output is often wrong or the boundary is ambiguous and blurred.
Therefore, we introduce a camouflage edge detection module that explicitly estimates the boundary while accounting for the uncertainty, and we call this new framework as \textbf{uncertainty-reduction COD} (UR-COD).
As shown in Figure~\ref{fig:our_framework}, our framework consists of three modules: pseudo-map generator~(PMG) (in Sec.\ref{sec31}), pseudo-edge generator~(PEG) (in Sec.\ref{sec32}), and uncertainty-aware map refinement~(UAMR) module (in Sec.\ref{sec33}).
The details of each module are described below.

\subsection{Pseudo-Map Generator~(PMG) \label{sec31}}
The output camouflaged map generated by the conventional COD methods often has unclear boundaries.
In this method, a coarse camouflaged map is generated from the RGB image $I$ using the conventional COD method, and it is treated as a pseudo-map label $M_{\rm pseudo}$.
In other words, PMG is a module that outputs a pseudo-map label $M_{\rm pseudo}$ from an RGB image $I$.
In this paper, we chose five methods as baselines for the COD models: SINet~\cite{fan2020camouflaged}, SINet-v2~\cite{fan2021concealed}, PraNet~\cite{fan2020pranet}, C$^2$FNet~\cite{sun2021context}, and MGL~\cite{zhai2021Mutual}, whose implementations are publicly available.
In PMG, any conventional method can be incorporated as a COD model, and the performance of our method is to improve the COD model to become a more powerful one.

\subsection{Pseudo-Edge Generator~(PEG) \label{sec32}}
PEG is a module that outputs a pseudo-edge label $E_{\rm pseudo}$ from an RGB image $I$.
To achieve this, we need to train some kind of edge detection model suitable for the camouflaged object datasets.
However, most of the objects in the datasets handled by existing edge detection models are salient objects, and they cannot perform well on camouflaged objects.
To deal with this problem and perform camouflaged edge detection, we follow CENet~\cite{gu2019net}, which extracts the context of the input image via DAC blocks~\cite{gu2019net} and RMP blocks~\cite{gu2019net}.
This is based on the hypothesis that, medical image analysis and COD are two very close tasks, in which the object in a medical image is also not salient \cite{fan2020pranet}.
Therefore, it is intuitive to apply CENet to the case where the object is concealed, which fully considers semantic information of images.

The DAC block contains four cascading branches that gradually increase the number of atrous convolutions, formulated as:
\begin{equation}
	y[i] = \sum_{k}x[i+rk]w[k],
\end{equation}
where the convolution of the input feature map $x$ and a filter $w$ yields the output $y$, and the atrous rate $r$ corresponds to the stride at which the input signal is sampled. 
Therefore, the DAC block can extract features from different scales.
In the RMP block, after gathering context information with four different sizes of pooling kernels, features are fed into 1$\times$1 convolution and combined with the original features.
Using these blocks, the semantic information of the context is extracted and we can achieve a high-level feature map.
We use binary cross-entropy (BCE) loss to train the PEG.
However, according to our observation through experiment, directly using the classical BCE loss will face the problem of overfitting.
To mitigate this, we introduce a new edge loss with Flooding~\cite{ishida2020we} that is calculated as:
\begin{equation}
	\mathcal{L}_{\rm edge} = \mid \mathcal{L}_{\rm bce}(E_{\rm pseudo}, E_{\rm GT}) - b\mid  + b,
\end{equation}
where $b$ is the flooding level, which sets the target value of loss to a small constant.

\subsection{Uncertainty-Aware Map Refinement Module~(UAMR) \label{sec33}}
To obtain the clear boundaries, we introduce a refinement module to refine the obscure boundaries of pseudo-map labels $M_{\rm pseudo}$ generated from PMG together using pseudo-edge labels $E_{\rm pseudo}$ generated from PEG.
However, these pseudo labels are predicted by the learning-based model, and thus contain uncertainty.
Inspired by UCNet~\cite{zhang2020uc}, we propose a UAMR module that can absorb uncertainty.
Similar to~\cite{zhang2020uc}, we assume that pseudo-map labels and pseudo-edge labels have similar properties as depth labels in that they provide clues for object detection. 
By extending UCNet to support pseudo-map labels and pseudo-edge labels, we construct the UAMR module that considers uncertainty.
Therefore, the input to the UAMR module is an RGB image $I$, a pseudo-map label $M_{\rm pseudo}$, and a pseudo-edge label $E_{\rm pseudo}$, which are used to predict an edge-accurate camouflaged map $M_{\rm pred}$.
As shown in Figure~\ref{fig:our_framework}, the UAMR module consists of PriorNet, PosteriorNet, and RefinementNet.

PriorNet takes the RGB image $I$, the pseudo-map label $M_{\rm pseudo}$, and the pseudo-edge label $E_{\rm pseudo}$ as input and maps them to a low-dimensional latent variable $z^{\rm prior}$.
On the other hand, PosteriorNet takes all the input from PriorNet together with the ground-truth map label $M_{\rm GT}$ as input and maps them to a low-dimensional latent variable $z^{\rm post}$.
Note that PriorNet and PosteriorNet form CVAE, and each latent variable $z^{\rm prior}, z^{\rm post}$ is sampled from the Gaussian distribution consisting of the parameters of the output mean $\mu^{\rm prior}, \mu^{\rm post}$ and variance $\sigma^{\rm prior}, \sigma^{\rm post}$.
As for the CVAE loss, let $X$ denotes $(I, M_{\rm pseudo}, E_{\rm pseudo})$, let $Y$ denotes the ground-truth map $M_{\rm GT}$.
The PriorNet is defined as $P_{\theta}(z^{\rm prior}\mid X)$ and the PosteriorNet is defined as $Q_{\phi}(z^{\rm post}\mid X,Y)$, where $\theta$ is the parameter set of PriorNet and $\phi$ is that of PosteriorNet.
The CVAE loss is defined as:
\begin{multline}
	\mathcal{L}_{\rm CVAE} = E_{z\sim Q_{\phi}(z^{\rm post}\mid X,Y)}[-\log P_{\omega}(Y\mid X,z^{\rm post})] \\ 
	+ D_{\rm KL}(Q_{\phi}(z^{\rm post}\mid X,Y)\parallel P_{\theta}(z^{\rm prior}\mid X)),
\end{multline}
where $P_{\omega}(Y\mid X,z^{\rm post})$ is the likelihood of $P(Y)$ given latent variable $z^{\rm post}$ and conditioning variable $X$, and $D_{\rm KL}$ is Kullback-Leibler Divergence.

RefinementNet takes the RGB image $I$, the pseudo-map label $M_{\rm pseudo}$, the pseudo-edge label $E_{\rm pseudo}$, and each latent variable $z^{\rm prior}, z^{\rm post}$ as input, which is trained to estimate the camouflaged map $M_{\rm pred}$, the corrected pseudo-edge label $E_{\rm ref}$, and the corrected pseudo-map label $M_{\rm ref}$.
Therefore, using the latent variables in CVAE, we can consider the uncertainty of the pseudo-map labels and pseudo-edge labels by learning to make the reconstructed pseudo-map label $M_{\rm ref}$ closer to the ground-truth $M_{\rm GT}$ and the reconstructed pseudo-edge label $E_{\rm ref}$ closer to the ground-truth edge label $E_{\rm GT}$.
As for the loss for RefinementNet, MSE loss $\mathcal{L}_{\rm mse}$, smoothness loss~\cite{godard2017unsupervised} $\mathcal{L}_{\rm smooth}$, and structure loss~\cite{luo2017non} $\mathcal{L}_{\rm struct}$ are used.
Smoothness loss is used to enhance the structure information in the image, while structure loss is used to enforce spatial coherence of the prediction and using both the local and global features in the optimization.
Thus, the refinement loss is defined as:
\begin{multline}
	\mathcal{L}_{\rm ref} = \lambda_{\rm mse}^{\rm prior}\mathcal{L}_{\rm mse}(P_{\rm ref}^{\rm prior}, P_{\rm GT}^{\rm prior})+\lambda_{\rm mse}^{\rm post}\mathcal{L}_{\rm mse}(P_{\rm ref}^{\rm post}, P_{\rm GT}^{\rm post}) \\
	+\lambda_{\rm smooth}^{\rm prior}\mathcal{L}_{\rm smooth}(M_{\rm pred}^{\rm prior}, M_{\rm GT})+\lambda_{\rm smooth}^{\rm post}\mathcal{L}_{\rm smooth}(M_{\rm pred}^{\rm post}, M_{\rm GT}) \\
	+\lambda_{\rm struct}^{\rm prior}\mathcal{L}_{\rm struct}(M_{\rm pred}^{\rm prior}, M_{\rm GT})+\lambda_{\rm struct}^{\rm post}\mathcal{L}_{\rm struct}(M_{\rm pred}^{\rm post}, M_{\rm GT}),
\end{multline}
where $P$ denotes $(M, E)$, and $\lambda$ denotes the weight for each loss.

Finally, combing the above three modules together, we get the overall loss for our model:
\begin{equation}
	\mathcal{L} = \lambda_{\rm edge}\mathcal{L}_{\rm edge}+\lambda_{\rm CVAE}\mathcal{L}_{\rm CVAE}+\lambda_{\rm ref}\mathcal{L}_{\rm ref},
\end{equation}
where $\lambda$ denotes the weight for each loss. 
In testing, the output camouflaged map $M_{\rm pred}^{\rm prior}$ of RefinementNet is treated as the final output of our framework.

\newcolumntype{A}{>{\centering\arraybackslash}p{15mm}}
\begin{table*}[tb]
\small
{\tabcolsep=0.8mm
\caption{Quantitative results on four standard COD datasets. $\uparrow$ indicates the higher the score the better, and vice versa.} \vspace{-1em}
\begin{center}
\begin{tabular}{l|cccc|cccc|cccc|cccc}
\toprule
\multirow{2}{*}{Methods} & \multicolumn{4}{c|}{CHAMELEON} & \multicolumn{4}{c|}{CAMO-Test} & \multicolumn{4}{c|}{COD10K-Test} & \multicolumn{4}{c}{NC4K} \\ \cline{2-17} 
                         & $S_\alpha\uparrow$ & $E_\phi\uparrow$ & $F_\beta^{w}\uparrow$ & $\mathcal{M}\downarrow$ & $S_\alpha\uparrow$ & $E_\phi\uparrow$ & $F_\beta^{w}\uparrow$ & $\mathcal{M}\downarrow$ & $S_\alpha\uparrow$ & $E_\phi\uparrow$ & $F_\beta^{w}\uparrow$ & $\mathcal{M}\downarrow$ & $S_\alpha\uparrow$ & $E_\phi\uparrow$ & $F_\beta^{w}\uparrow$ & $\mathcal{M}\downarrow$ \\ \hline
FPN~\cite{lin2017feature} & 0.794  & 0.783 & 0.590 & 0.075 & 0.684  & 0.677 & 0.483 & 0.131 & 0.697  & 0.691  & 0.411 & 0.075 & -     & -      & -     & -    \\
MaskRCNN~\cite{He_2017_ICCV} & 0.643  & 0.778 & 0.518 & 0.099 & 0.574  & 0.715 & 0.430 & 0.151 & 0.613  & 0.748  & 0.402 & 0.080 & -     & -      & -     & -    \\
PSPNet~\cite{zhao2017pyramid} & 0.773  & 0.758 & 0.555 & 0.085 & 0.663  & 0.659 & 0.455 & 0.139 & 0.678  & 0.680  & 0.377 & 0.080 & -     & -      & -     & -    \\
UNet++~\cite{zhou2018unet++} & 0.695  & 0.762 & 0.501 & 0.094 & 0.599  & 0.653 & 0.392 & 0.149 & 0.623  & 0.672  & 0.350 & 0.086 & -     & -      & -     & -    \\
PiCANet~\cite{liu2018picanet} & 0.769  & 0.749 & 0.536 & 0.085 & 0.609  & 0.584 & 0.356 & 0.156 & 0.649  & 0.643  & 0.322 & 0.090 & -     & -      & -     & -    \\
MSRCNN~\cite{huang2019mask} & 0.637  & 0.686 & 0.443 & 0.091 & 0.617  & 0.669 & 0.454 & 0.133 & 0.641  & 0.706  & 0.419 & 0.073 & -     & -      & -     & -    \\
PFANet~\cite{zhao2019pyramid} & 0.679  & 0.648 & 0.378 & 0.144 & 0.659  & 0.622 & 0.391 & 0.172 & 0.636  & 0.618  & 0.286 & 0.128 & -     & -      & -     & -    \\
CPD~\cite{wu2019cascaded} & 0.853  & 0.866 & 0.706 & 0.052 & 0.726  & 0.729 & 0.550 & 0.115 & 0.747  & 0.770  & 0.508 & 0.059 & -     & -      & -     & -    \\
HTC~\cite{chen2019hybrid} & 0.517  & 0.489 & 0.204 & 0.129 & 0.476  & 0.442 & 0.174 & 0.172 & 0.548  & 0.520  & 0.221 & 0.088 & -     & -      & -     & -    \\
ANet-SRM~\cite{le2019anabranch} & - & - & - & - & 0.682  & 0.685 & 0.484 & 0.126 & - & - & - & - & -     & -      & -     & -    \\
EGNet~\cite{zhao2019egnet} & 0.848  & 0.870 & 0.702 & 0.050 & 0.732  & 0.768 & 0.583 & 0.104 & 0.737  & 0.779  & 0.509 & 0.056 & -     & -      & -     & -    \\
MirrorNet~\cite{yan2020mirrornet} & - & - & - & - & 0.741  & 0.804 & 0.652 & 0.100 & - & - & - & - & -     & -      & -     & -    \\
TIGNet~\cite{zhu2021inferring} & 0.874 & 0.916 & 0.783 & 0.038 & 0.781  & 0.847 & 0.678 & 0.087 & 0.793 & 0.848 & 0.635 & 0.043 & -     & -      & -     & -    \\
PFNet~\cite{mei2021camouflaged} & 0.882 & - & 0.810 & 0.033 & 0.782  & - & 0.695 & 0.085 & 0.800 & - & 0.660 & 0.040 & -     & -      & -     & -    \\
LSR~\cite{yunqiu_cod21} & 0.893 & 0.938 & - & 0.033 & 0.793 & 0.826 & - & 0.085 & 0.793 & 0.868 & - & 0.041 & 0.839 & 0.883 & - & 0.053 \\\hline
SINet~\cite{fan2020camouflaged} & 0.872  & 0.936 & 0.806 & 0.034 & 0.745  & 0.804 & 0.644 & 0.092 & 0.776  & 0.864  & 0.631 & 0.043 & 0.808 & 0.871 & 0.723 & 0.058 \\
UR-SINet~(Ours) & 0.876  & 0.942 & 0.824 & 0.031 & 0.741  & 0.804 & 0.649 & 0.091 & 0.775  & 0.869  & 0.643 & 0.041 & 0.806 & 0.873 & 0.731 & 0.057 \\\hline
PraNet~\cite{fan2020pranet} & 0.860  & 0.907 & 0.763 & 0.044 & 0.769  & 0.824 & 0.663 & 0.094 & 0.789  & 0.861  & 0.629 & 0.045 & 0.822 & 0.876 & 0.724 & 0.059 \\
UR-PraNet~(Ours) & 0.884  & 0.936 & 0.835 & 0.032 & 0.762  & 0.831 & 0.679 & 0.089 & 0.790  & 0.877  & 0.661 & 0.039 & 0.820 & 0.885 & 0.747 & 0.054 \\\hline
C$^2$FNet~\cite{sun2021context} & 0.888  & 0.935 & 0.828 & 0.032 & 0.796  & 0.854 & 0.719 & 0.080 & 0.813  & 0.890  & 0.686 & 0.036 & 0.839 & 0.897 & 0.766 & 0.049 \\
UR-C$^2$FNet~(Ours) & 0.889  & 0.940 & 0.844 & 0.029 & 0.791  & 0.856 & 0.725 & 0.079 & 0.811  & 0.897  & 0.700 & 0.034 & 0.836 & 0.900 & 0.775 & 0.047 \\\hline
MGL~\cite{zhai2021Mutual} & 0.892  & 0.913 & 0.802 & 0.032 & 0.772  & 0.807 & 0.664 & 0.089 & 0.811  & 0.844  & 0.654 & 0.037 & 0.829 & 0.863 & 0.730 & 0.055 \\
UR-MGL~(Ours) & 0.891  & 0.942 & 0.844 & 0.026 & 0.763  & 0.824 & 0.682 & 0.086 & 0.803  & 0.879  & 0.685 & 0.034 & 0.821 & 0.882 & 0.752 & 0.051 \\\hline
SINet-v2~\cite{fan2021concealed} & 0.888  & 0.942 & 0.816 & 0.030 & {\bf 0.820}  & 0.882 & 0.743 & 0.070 & 0.815  & 0.887  & 0.680 & 0.037 & {\bf 0.847} & 0.903 & 0.770 & 0.048 \\
UR-SINet-v2~(Ours) & {\bf 0.901}  & {\bf 0.960} & {\bf 0.862} & {\bf 0.023} & 0.814  & {\bf 0.891} & {\bf 0.758} & {\bf 0.067} & {\bf 0.816}  & {\bf 0.903}  & {\bf 0.708} & {\bf 0.033} & 0.844 & {\bf 0.910} & {\bf 0.787} & {\bf 0.045} \\\toprule

\end{tabular}
\end{center}
\label{table:quantitative_evaluation}
}
\end{table*}

\begin{table*}[tb]
\small
{\tabcolsep=0.8mm
\caption{Ablation experiments of the proposed model.} \vspace{-1em}
\begin{center}
\begin{tabular}{l|cccc|cccc|cccc|cccc}
\toprule
\multirow{2}{*}{Methods} & \multicolumn{4}{c|}{CHAMELEON} & \multicolumn{4}{c|}{CAMO-Test} & \multicolumn{4}{c|}{COD10K-Test} & \multicolumn{4}{c}{NC4K} \\ \cline{2-17} 
                         & $S_\alpha\uparrow$ & $E_\phi\uparrow$ & $F_\beta^{w}\uparrow$ & $\mathcal{M}\downarrow$ & $S_\alpha\uparrow$ & $E_\phi\uparrow$ & $F_\beta^{w}\uparrow$ & $\mathcal{M}\downarrow$ & $S_\alpha\uparrow$ & $E_\phi\uparrow$ & $F_\beta^{w}\uparrow$ & $\mathcal{M}\downarrow$ & $S_\alpha\uparrow$ & $E_\phi\uparrow$ & $F_\beta^{w}\uparrow$ & $\mathcal{M}\downarrow$ \\ \hline
SINet-v2~\cite{fan2021concealed} & 0.888  & 0.942 & 0.816 & 0.030 & {\bf 0.820}  & 0.882 & 0.743 & 0.070 & 0.815  & 0.887  & 0.680 & 0.037 & {\bf 0.847} & 0.903 & 0.770 & 0.048 \\
UR-SINet-v2 w/o PEG & 0.898  & 0.957 & 0.857 & 0.024 & 0.813  & 0.890 & 0.756 & 0.068 & 0.814  & 0.902  & 0.704 & 0.034 & 0.842 & 0.909 & 0.784 & 0.046 \\
UR-SINet-v2 w/o PMG & 0.873  & 0.919 & 0.814 & 0.038 & 0.705  & 0.737 & 0.571 & 0.105 & 0.742  & 0.805  & 0.573 & 0.047 & 0.786 & 0.841 & 0.687 & 0.063 \\
UR-SINet-v2~(Ours) & {\bf 0.901}  & {\bf 0.960} & {\bf 0.862} & {\bf 0.023} & 0.814  & {\bf 0.891} & {\bf 0.758} & {\bf 0.067} & {\bf 0.816}  & {\bf 0.903}  & {\bf 0.708} & {\bf 0.033} & 0.844 & {\bf 0.910} & {\bf 0.787} & {\bf 0.045} \\
\toprule
\end{tabular}
\end{center}
\label{table:ablation_study}
}
\end{table*}

\begin{figure*}[htbp]
  \begin{minipage}[b]{0.12\hsize}
    \centering
    \includegraphics[width=\hsize]{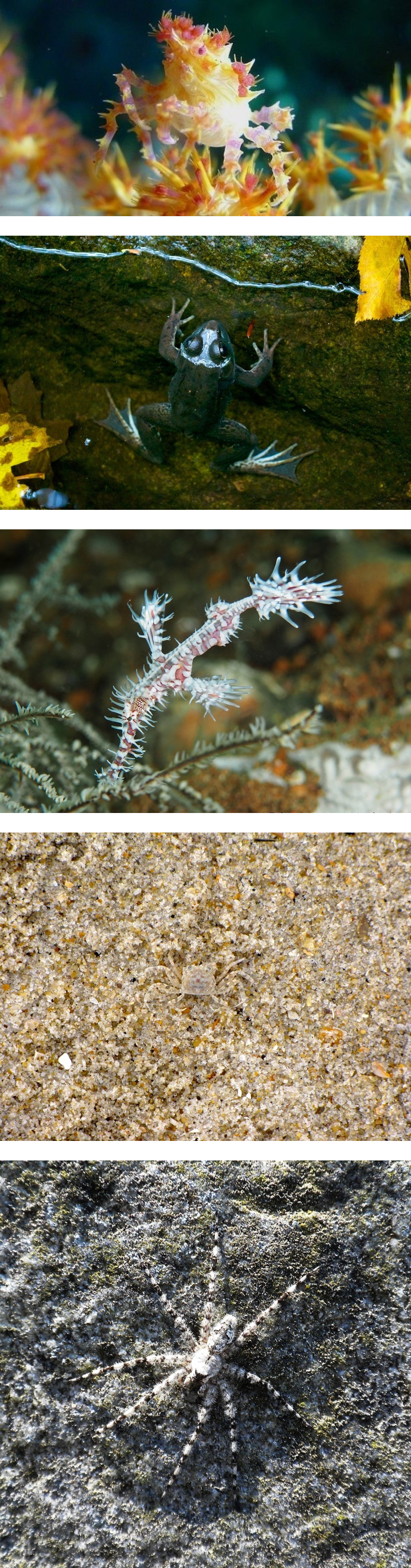}
    \subcaption{Image}
  \end{minipage}
  \begin{minipage}[b]{0.12\hsize}
    \centering
    \includegraphics[width=\hsize]{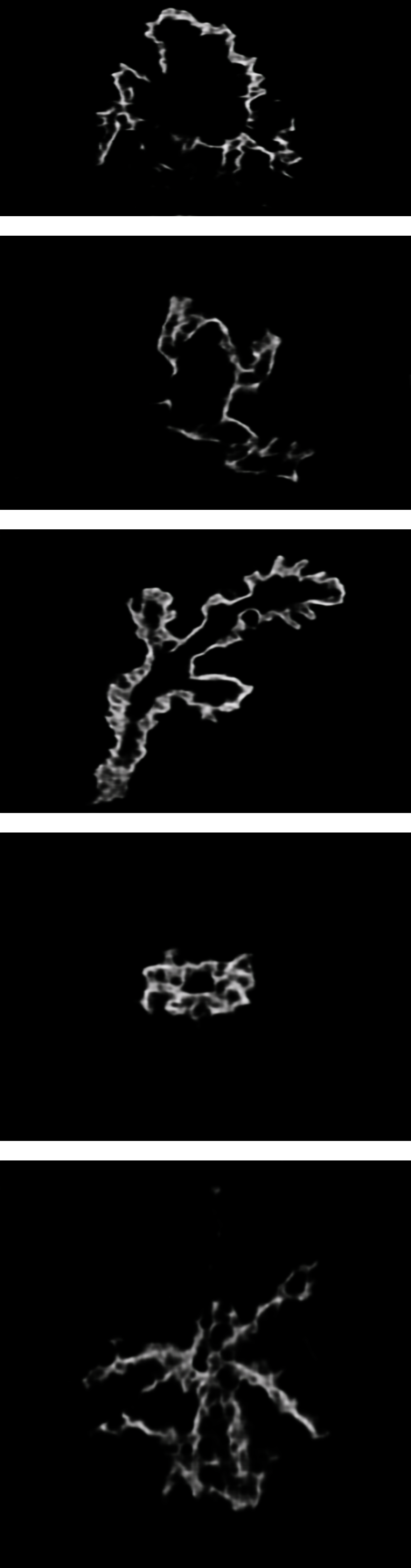}
    \subcaption{Pseudo-edge}
  \end{minipage}
  \begin{minipage}[b]{0.12\hsize}
    \centering
    \includegraphics[width=\hsize]{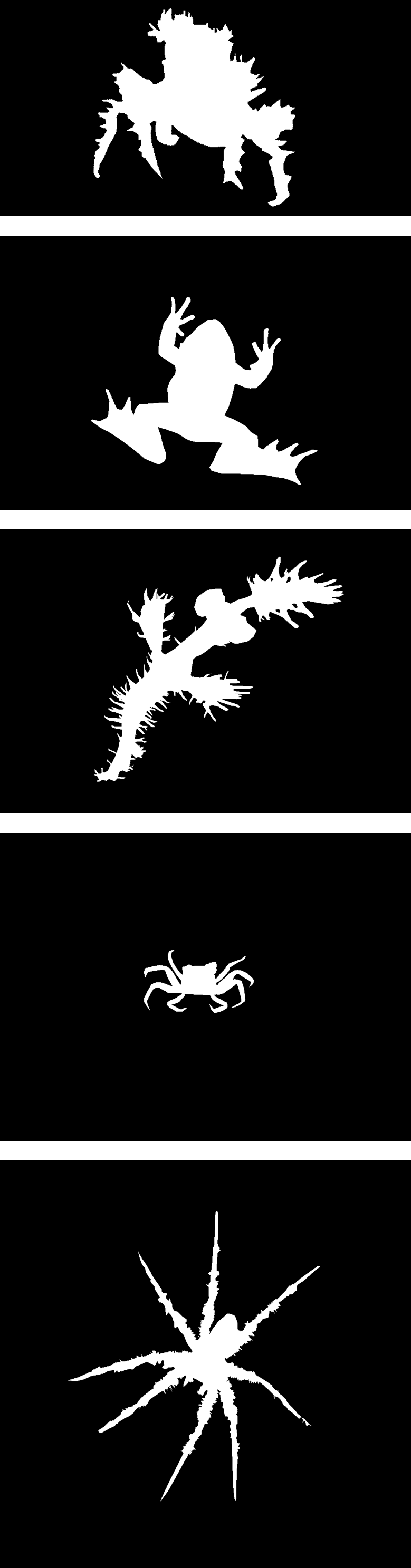}
    \subcaption{GT}
  \end{minipage}
  \begin{minipage}[b]{0.12\hsize}
    \centering
    \includegraphics[width=\hsize]{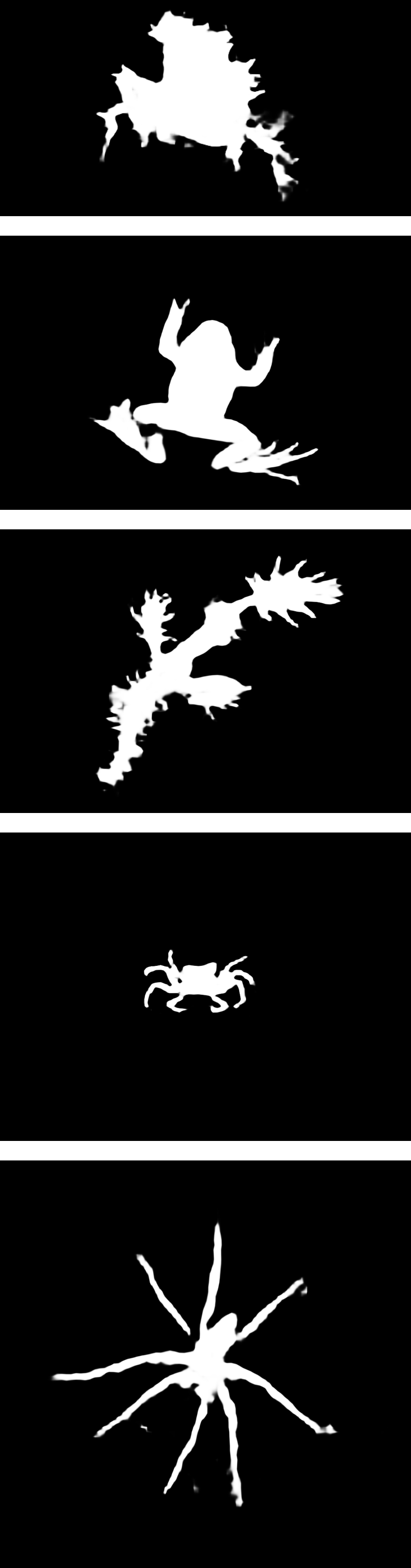}
    \subcaption{Ours}
  \end{minipage}
  \begin{minipage}[b]{0.12\hsize}
    \centering
    \includegraphics[width=\hsize]{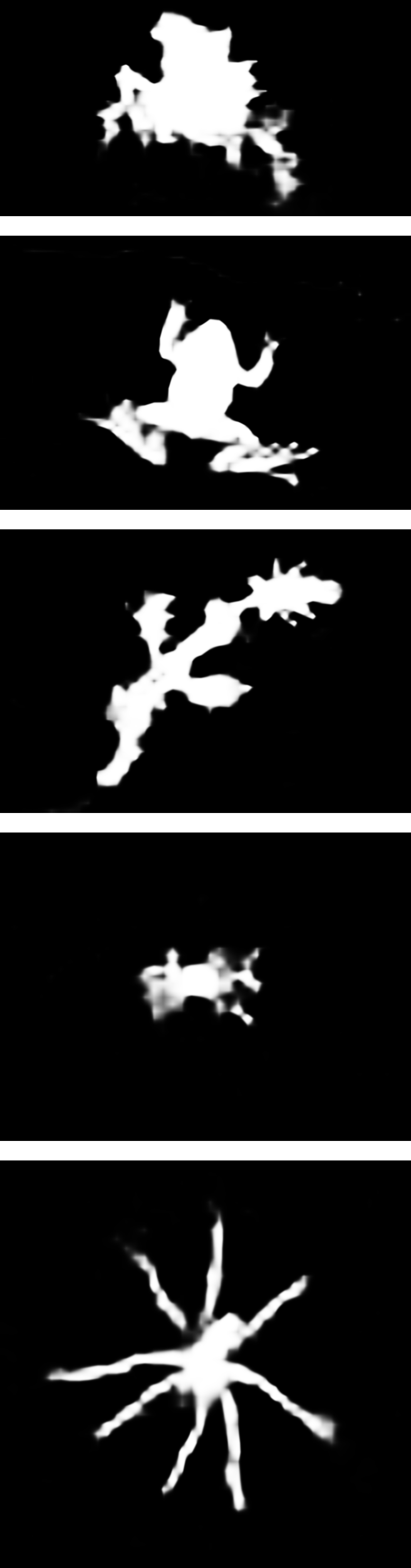}
    \subcaption{SINet-v2~\cite{fan2021concealed}}
  \end{minipage}
  \begin{minipage}[b]{0.12\hsize}
    \centering
    \includegraphics[width=\hsize]{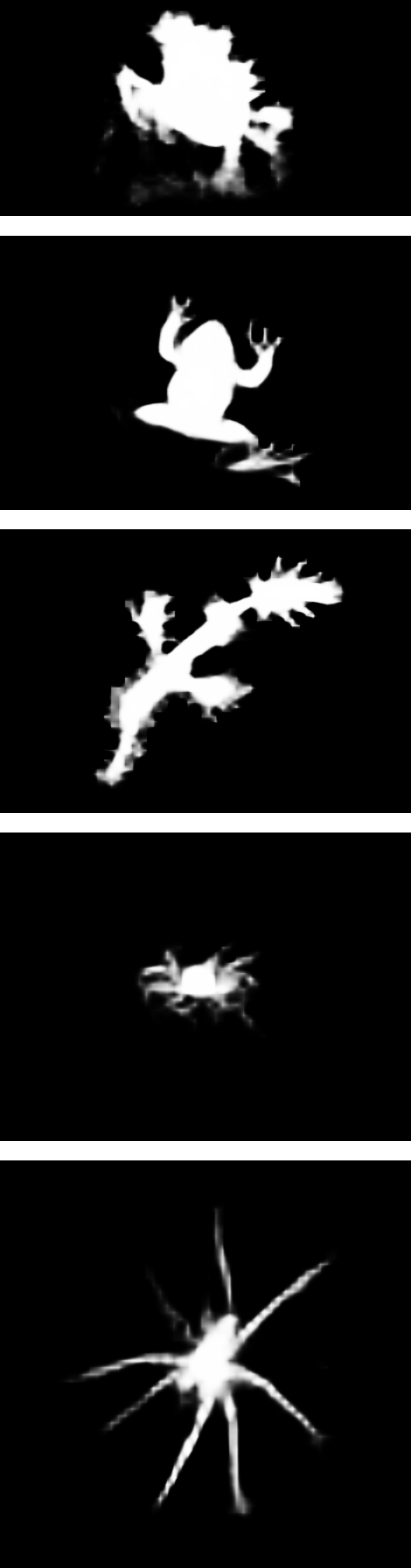}
    \subcaption{MGL~\cite{zhai2021Mutual}}
  \end{minipage}
  \begin{minipage}[b]{0.12\hsize}
    \centering
    \includegraphics[width=\hsize]{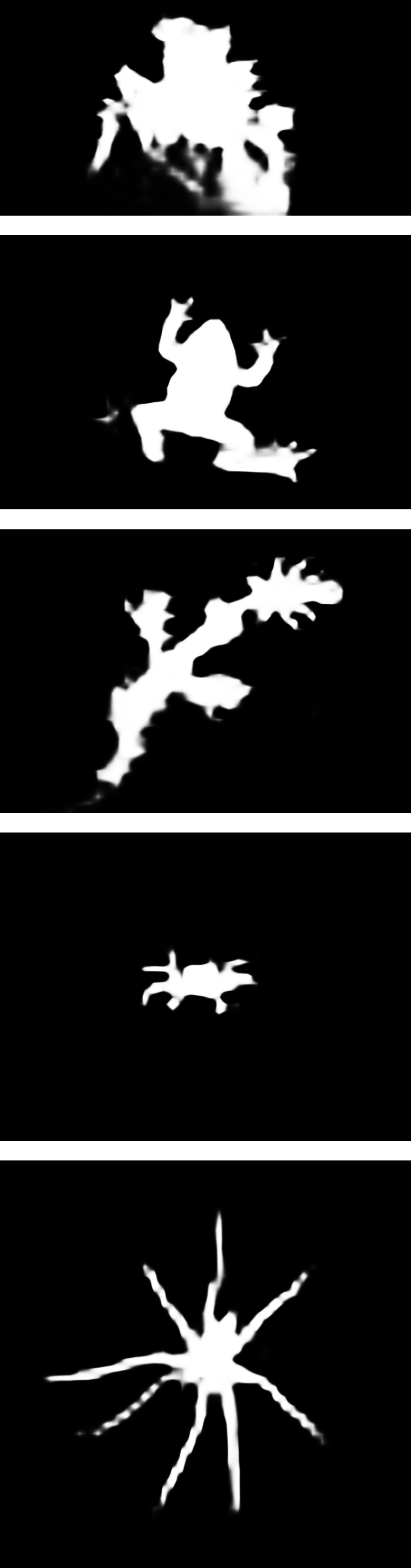}
    \subcaption{C$^2$FNet~\cite{sun2021context}}
  \end{minipage}
  \begin{minipage}[b]{0.12\hsize}
    \centering
    \includegraphics[width=\hsize]{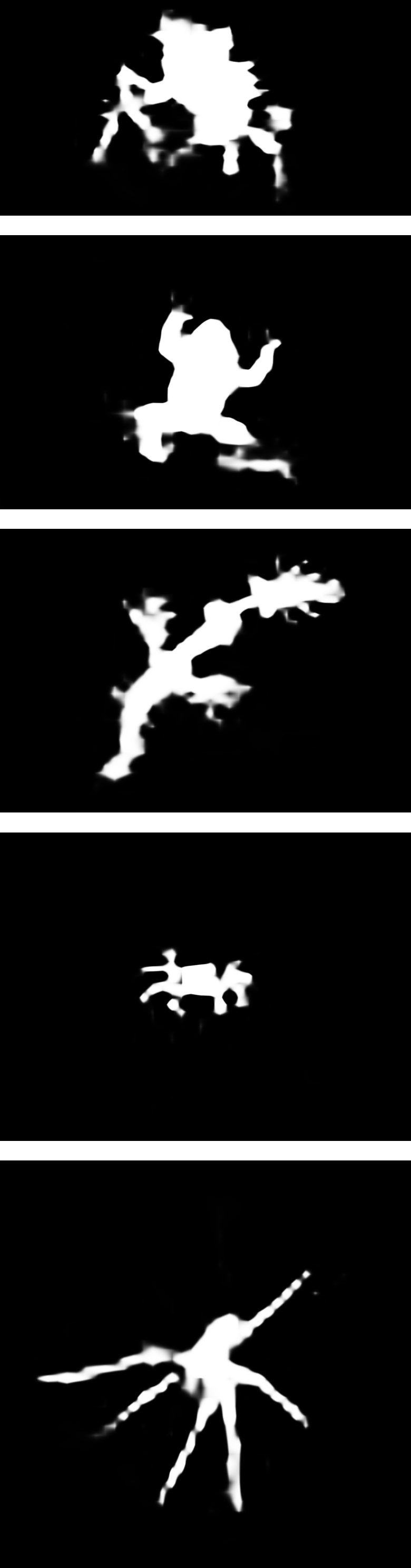}
    \subcaption{SINet~\cite{fan2020camouflaged}}
  \end{minipage}
  \caption{Visual comparison of the proposed model with state-of-the-art methods. "UR-SINet-v2" model is adopted as our model (d), and (e) is also used as a pseudo-map label for training our model. Our model outputs camouflaged maps with accurate boundaries based on the uncertain pseudo-map labels and pseudo-edge labels.}
  \label{fig:qualitative}
\end{figure*}

\section{Experiments}

\subsection{Experimental Settings}
\subsubsection{Implementation Details}
We implement our model using PyTorch, and initialized the encoder of RefinementNet with ResNet-50~\cite{he2016deep} parameters pre-trained on ImageNet and the encoder of PEG with the parameters of ResNet-34~\cite{he2016deep}. 
We resize all the images to 352$\times$352 for both training and testing.
The scale of latent space is set to 3, and the maximum epoch is 100.
We optimize the overall parameters using the Adam algorithm, where the initial learning rate is 5e-5 and after 80 epochs, the learning rate is reduced by $10\%$ for each epoch.
The whole training takes 8.5 hours with batch size 10 on an NVIDIA Tesla V100 GPU.
Besides, our approach has a low computational cost during the inference.

\subsubsection{Datasets}
To train our framework, we used a standard training dataset for COD which contains 3,040 images from the COD10K dataset~\cite{fan2020camouflaged} and 1,000 images from CAMO dataset~\cite{le2019anabranch}.
During training, we generated the ground-truth edge labels from the ground-truth map labels.
To evaluate our framework, we used the CAMO dataset consisting of 250 images with camouflaged objects, the CHAMELEON dataset~\cite{skurowski2018animal} consisting of 76 images, the COD10K dataset consisting of 2,026 images, and the NC4K dataset~\cite{yunqiu_cod21} consisting of 4,121 images.

\subsubsection{Evaluation Metrics}
Four quantitative evaluation metrics are widely used to evaluate the performance of CODs: Mean Absolute Error~(MAE), S-measure~\cite{fan2017structure}, E-measure~\cite{fan2018enhanced}, and weighted F-measure~\cite{margolin2014evaluate} denoted as $\mathcal{M}, S_{\alpha}, E_{\phi}$, and $F_{\beta}^{w}$, respectively.

$\mathcal{M}$ is defined as per-pixel-wise difference between the predicted map $M_{\rm pred}$ and the ground-truth map $M_{\rm GT}$ as: $\mathcal{M} = \frac{1}{H\times W}\mid M_{\rm pred}-M_{\rm GT}\mid$, where $H$ and $W$ are the height and width of $M_{\rm pred}$.
The MAE directly evaluates the conformity between the estimated map and the ground-truth map.
S-measure $S_{\alpha}$ is a structure-based metric that combines region-aware structural similarity $S_{r}$ and object-aware structural similarity $S_{o}$ as: $S_{\alpha}=\alpha S_o+(1-\alpha)S_r$, and the balance parameter $\alpha$ is set to 0.5 as default.
E-measure $E_{\phi}$ simultaneously evaluates the local pixel-level matching information and the image-level statistics.
Weighted F-measure $F_{\beta}^{w}$ defines a weighted precision that can provide more reliable evaluation results than F-measure, which is a comprehensive measure of both precision and recall of the predicted camouflaged map.

\subsection{Comparison with State-of-the-arts}
We compare our method against 20 state-of-the-art baselines: object detection method FPN~\cite{lin2017feature}; semantic segmentation method PSPNet~\cite{zhao2017pyramid}; instance segmentation methods MaskRCNN~\cite{He_2017_ICCV}, HTC~\cite{chen2019hybrid}, and MSRCNN~\cite{huang2019mask}; medical image segmentation methods UNet++~\cite{zhou2018unet++} and PraNet~\cite{fan2020pranet}; salient object detection methods PiCANet~\cite{liu2018picanet}, CPD~\cite{wu2019cascaded}, PFANet~\cite{zhao2019pyramid}, EGNet~\cite{zhao2019egnet}; and COD methods ANet-SRM~\cite{le2019anabranch}, SINet~\cite{fan2020camouflaged}, MirrorNet~\cite{yan2020mirrornet}, C$^2$FNet~\cite{sun2021context}, TIGNet~\cite{zhu2021inferring}, PFNet~\cite{mei2021camouflaged}, MGL~\cite{zhai2021Mutual}, LSR~\cite{yunqiu_cod21}, SINet-v2~\cite{fan2021concealed}.
For a fair comparison, the results of the non-COD methods are taken from~\cite{fan2021concealed}, and the results of the COD methods are taken from the respective papers, some of which are obtained by output camouflaged maps provided on public websites or running models retrained with open source code. 

\subsection{Quantitative Evaluation}
Table~\ref{table:quantitative_evaluation} shows the metric scores of the proposed method and 20 state-of-the-art baselines on the four benchmark datasets.
We can see that our method of "UR-SINet-v2~(Ours)" outperforms almost all the other methods on four standard metrics: S-measure, E-measure, weighted F-measure, and MAE, which achieves average improvements of $0.13\%$, $3.5\%$, $1.4\%$, and $11.2\%$, respectively, when compared to the SINet-v2.
Moreover, although the performance of the S-measure is slightly lower, the overall scores of the other metrics are significantly improved compared to the original COD methods without our framework, due to considering the uncertainty of pseudo-edge labels and pseudo-map labels.
The reason why the score of S-measure is slightly low is that the output camouflaged maps of our method have clear boundaries, and the penalty for failing to predict the true camouflaged maps is much larger for S-measure, which measures structural similarity, than for the case where the boundaries are ambiguous.

\subsection{Visualization}
Figure~\ref{fig:qualitative} shows the visual examples that are generated by our method and compared methods.
We can see that the camouflaged map is refined by appropriately combining the information of the pseudo-map label and the pseudo-edge label.
In particular, details such as tactile sensation and limbs of camouflaged objects, which tend to be difficult to detect by conventional methods, can be estimated accurately by the proposed method.

\subsection{Ablation Study}
\subsubsection{Effectiveness of Pseudo-Edge Labels}
In the baseline setting, we generate pseudo-edge labels from PEG and pseudo-map labels from PMG, then input both pseudo-edge and pseudo-map labels into the UAMR module and obtain the output camouflaged maps.
Table~\ref{table:ablation_study} shows the result of not inputting the pseudo-edge labels into the UAMR module, and the performance gets slightly lower.
In our framework, the edge information helps to correct the edges of the camouflaged maps and make the details of the edges better.
However, the metric scores focus on analyzing the scores of the overall effects.
So, the visualization is better but the increase in scores is not obvious.

\subsubsection{Effectiveness of Pseudo-Map Labels}
In contrast, table~\ref{table:ablation_study} also shows the result of not inputting the pseudo-map labels into the UAMR module, and the performance is considerably lower.
This is because pseudo-edge labels alone may not capture all the edges of camouflaged objects, and they contain a relatively large amount of uncertainty.

\section{Conclusions}
In this study, we dealt with camouflaged object detection, which is a challenging task to detect objects hidden in the environment.
Previous COD models, that directly segment objects, often result in blurred and ambiguous boundaries of the output camouflaged map, while models that consider edge information have slightly lower performance.
To handle these problems, we aim to combine the best of both methods and propose to refine the pseudo-map labels generated from conventional COD methods by referring to the pseudo-edge labels generated from a camouflaged edge detection module.
To solve the problem that pseudo-map labels and pseudo-edge labels are noisy and contain uncertainty, we proposed an uncertainty-aware map refinement module that outputs edge-accurate camouflaged maps.
We conducted a quantitative evaluation on the standard four COD datasets and found that the proposed method outperformed the state-of-the-art methods in almost all of the four evaluation metrics.

\bibliographystyle{ACM-Reference-Format}
\bibliography{reference}

\end{document}